\documentclass{article}

    \PassOptionsToPackage{numbers, compress}{natbib}
    \usepackage[preprint]{neurips_2020}

\usepackage{xcolor}
\usepackage[letterpaper]{geometry}
\usepackage[utf8]{inputenc} % allow utf-8 input
\usepackage[T1]{fontenc}    % use 8-bit T1 fonts
\usepackage{hyperref}       % hyperlinks
\usepackage{url}            % simple URL typesetting
\usepackage{booktabs}       % professional-quality tables
\usepackage{amsfonts}       % blackboard math symbols
\usepackage{nicefrac}       % compact symbols for 1/2, etc.
\usepackage{microtype}      % microtypography
\usepackage{amsmath}
\usepackage[ruled,vlined]{algorithm2e}
\usepackage[pdftex]{graphicx}

\title{Adversarial Optimal Transport Through The Convolution Of Kernels With Evolving Measures}

\author{%
  Daeyoung Kim \\
  Department of Mathematics\\
  New York University\\
  New York, NY 10012\\
  \texttt{kimdy@cims.nyu.edu} \\
   \And
   Esteban G. Tabak \\
   Department of Mathematics\\
   New York University\\
   New York, NY 10012\\
   \texttt{tabak@cims.nyu.edu} \\
}
\newcommand{\E}{\mathbb{E}}
\newcommand{\mb}{\mathbb}
\newcommand{\al}[1]{\begin{align*}{#1}\end{align*}}

\DeclareMathOperator*{\argmin}{arg\,min}

\begin{document}

\maketitle

% PAPER TOPIC:
% *primary*
% Adversarial Learning 
% *secondary*
% Adaptive Data Analysis 
% Density Estimation

% \begin{itemize}
%     \item Write acknowledgement
%     \item Review Style Rules
% \end{itemize}

\begin{abstract}
A novel algorithm is proposed to solve the sample-based optimal transport problem. An adversarial formulation of the push-forward condition uses a test function built as a convolution between an adaptive kernel and an evolving probability distribution $\nu$ over a latent variable $b$. Approximating this convolution by its simulation over evolving samples $b^i(t)$ of $\nu$, the parameterization of the test function reduces to determining the flow of these samples.  This flow, discretized over discrete time steps $t_n$, is built from the composition of elementary maps. The optimal transport also follows a flow that, by duality, must follow the gradient of the test function. The representation of the test function as the Monte Carlo simulation of a distribution makes the algorithm robust to dimensionality, and its evolution under a memory-less flow produces rich, complex maps from simple parametric transformations. The algorithm is illustrated with numerical examples.
\end{abstract}

% 1. how test function is constructed integral of kernel fun
% 2. 
% 3. 
\section{Introduction}
The optimal transport problem (OT), first proposed by Monge \citep{monge1781memoire} in 1781, concerns the movement of mass from $x\sim\rho$ to $y\sim\mu$ that minimizes a total transportation cost $\mb{E}[c(x,y)]$, where $x,y\in\Omega\subset\mb{R}^d$ and $\rho,\mu\in P(\Omega)$ are probability distributions with finite second moments. 
For ease of notation, we will denote both the distributions and their densities by $\rho$ or $\mu$.
Our focus is to solve the problem without a complete knowledge of the distributions, which are only revealed through samples. 
There has been much recent attention to OT in the machine learning community, especially from a density estimation perspective and as a natural tool for generative models. 
The transportation cost, which can be interpreted as a measure of the data distortion by the map, also provides a useful notion of distance among distributions. 
Monge's formulation of the problem seeks the cost minimizing map $T:\Omega\subset\mb{R}^d\to \mb{R}^d$ among those that push forward $\rho$ onto $\mu$, i.e. $T\#\rho = \mu$:
\[
T^* = \argmin_{T\#\rho=\mu} \int_\Omega c(x,T(x)) \rho(x) dx
\]
With the distributions known only through samples, it is not obvious a priori how to enforce the push-forward condition. 
We describe how we utilize an adversarial formulation over suitable function classes to efficiently estimate the optimizer $T$ in section \ref{section:methods}. 
In section \ref{section:examples}, we illustrate the algorithm with numerical examples in density estimation and simulation, and discuss future work and concluding remarks in section \ref{section:conclusion}.

\paragraph{Contributions}
We introduce a novel algorithm for the adversarial formulation of sample-based optimal transport problem that
\begin{itemize}
    \item can economically approximate the test function as a convolution between kernels and evolving measures; 
    \item can build rich, complex maps from elementary maps in a memory-less fashion;
    \item can serve as a general tool for density estimation and a generative modeling;
    \item is robust to space dimensionality.
\end{itemize}
\section{Related works}
\label{section:relatedworks}
Much of the previous work on OT, both theoretical and numerical, focused on solving the problem between known distributions, whether continuous or discrete. 
For instance \cite{cuturi2013sinkhorn} solved an entropy-regularized Kantorovich formulation of the OT problem between two discrete distributions. 
Many PDE-numerical studies of the problem have focused on spatial discretization \cite{chartrand2009gradient, benamou2000computational, haber2010efficient, haker2004optimal, iollo2011lagrangian}. 
By contrast, methods for sample-based OT were discussed in \cite{kuang2019sample, tabak2018conditional, tabak2018explanation}, with the push-forward condition relaxed to matching predetermined features over two sample sets. 
The resulting challenge in feature selection was addressed in \cite{essid2019adaptive}, which proposed an adversarial formulation of the OT problem through a variational characterization of the Kullback–Leibler divergence $\left(D_{KL}(\cdot\Vert\cdot)\right)$ in order to develop an adaptive feature selection method.
In \cite{tabak2019data}, a similar idea was extended to a conditional density estimation problem through the Wasserstein Barycenter problem \cite{agueh2011barycenters}. 

Our work is inspired by normalizing flows \cite{tabak2010density}, which introduced the power of composition of elementary maps and applications to density estimation via the change of variables formula 
\[
\rho(x) = \vert\det{\nabla_x T(x)}\vert\mu(T(x).
\]
% \textcolor{red}{NEED TO TALK ABOUT NORMALIZING FLOW FORMULA HERE.}
Similarly to previous flow-based solutions to the OT problem \cite{kuang2019sample, essid2019adaptive, tabak2019data}, we use the composition of elementary maps to approximate a function of arbitrary complexity. 
Our algorithm utilizes preconditioning, as developed in \cite{kuang2017preconditioning}, and the game-theory inspired implicit gradient method (IG) \cite{essid2019implicit} for the solution of minimax problems. We added a little twist to the optimization method, enforcing a trust region condition during the search at each iteration.

\section{Adversarial optimal transport}
\label{section:methods}
This section elaborates on the specifics of the problem and proposes the new method in detail.
\paragraph{Adversarial formulation}
Given $\Omega\subset \mb{R}^d$, $\rho, \mu \in P(\Omega)$, and a cost function $c(x,y)$, we address the optimal transport problem in Monge's formulation: 
\[
J = \min_{T\#\rho=\mu} \int_\Omega c(x, T(x)) \rho(x) dx = \E[c(X,T(X))]
\]
A weak formulation of the push-forward condition $T\#\rho=\mu$ is that for all test functions $F$, the condition $\E_{X\sim\rho}\left[F(T(X))\right]=\E_{Y\sim\mu}\left[F(Y)\right]$ must be met, so we have
\begin{eqnarray*}
J &=& \min_{T} \E[c(X,T(X))] \quad \text{s.t.} \quad  \forall F\quad \E[F(T(X))]=\E[F(Y)] \\
  &=& \min_{T} \max_{F} \E[c(X,T(X))] + \E[F(T(X))] - \E[F(Y)],
\end{eqnarray*}
providing an adversarial formulation of the problem. Because $\rho$ and $\mu$ appear only in the calculation of the expected value of functions, it is straightforward to obtain a sample-based formulation: given the sample sets $\{x_1,\ldots,x_N\} \sim \rho$ and $\{y_1,\ldots,y_M\} \sim \mu$,
\begin{eqnarray*}
\Tilde{J} = \min_{T} \max_{F} \frac{1}{N}\sum_{i=1}^N c(x_i,T(x_i)) + \frac{1}{N}\sum_{i=1}^N F(T(x_i)) - \frac{1}{M}\sum_{j=1}^M F(y_j) .
\end{eqnarray*}
Throughout this paper, we work with the $L^2$-norm cost: $c(x,y)=\Vert x-y \Vert^2/2$, and refer to the cost term in $J$ as \textit{cost} and the difference between the other two terms as \textit{constraint}.

\paragraph{Flow-based functions}
In order to complete the formulation above, we need to specify over which functional spaces for the map $T$ and the test function $F$ to perform the optimization. Ideally, the family of test functions should be rich enough for the objective function to blow up if $T\# \rho$ and $\mu$ should not match, yet not so rich as to detect differences attributable to the finite number of samples provided. Similarly, the family of maps should be able to push forward one distribution onto the other without overfitting the data points, for instance by moving each point independently. 

It has been proposed in \cite{tabak2010density} to use, in lieu of a global map $y=T(x)$, a flow $z(x, t)$, such that $z(x, 0) = x$ and $\lim_{t\rightarrow \infty} z(x, t) = T(x)$. A continuous flow discretizes naturally into map composition, with a number of advantages:
\begin{enumerate}
    \item Complex maps $T$ can be built through the composition of simpler, elementary maps $E_n$, each depending on only a handful of parameters.
    \item Rather than specifying a functional space for $T$, one can much more easily impose desired features, such as smoothness, on each $E_n$.
    \item By performing a memory-less optimization, i.e. only over the parameters of the current elementary map $E_n$, the computational complexity of each step is highly reduced.
\end{enumerate}
The main contribution of this article is to extend the parameterization through flows to the test function $F$. Unlike $T$, which maps $\mb{R}^n$ to itself, $F$ maps $\mb{R}^n$ to $\mb{R}$, so $F$ itself cannot really flow. Instead, we describe $F$ as a convolution between a kernel function and a probability density, represent the latter through sample ponts $\{b^k\}$, the ``representers'', and let these flow through the composition of elementary maps. This representation has the additional advantage of reducing the curse of dimensionality, as the error incurred through the Monte Carlo simulation of an integral --our convolution-- scales with the number of sample points, not the dimension of the space.

The elementary maps available at each step to evolve $T$ and $F$ must be related, as the two constitute the game's adversarial strategies: it would not make sense, for instance, for $T$ to be able to perform local movements in some domain while $F$ enforces the push-forward condition elsewhere. It turns out that duality considerations inform the relationship between the two. The resulting algorithm is detailed in the subsections below.

\subsection{Warm-up with fixed features}
Let us consider first a simple, feature-based approach to modeling the test functions, which then a flow-based approach will generalize. Given $K$ feature functions $\phi_j(y)$, such as, in one dimension, the monomials $\phi_j(y) = y^j$, define
\begin{eqnarray*}
F(y; \beta) = \sum_{j=1}^K \beta_j \phi_j(y) = \langle \beta, \phi(y) \rangle,
\end{eqnarray*}
where $\beta\in\mb{R}^K$ is a parameter, the ``representer'' of $F$.
For the transport map, we propose a flow discretized into
the composition of many elementary maps of the form
\[
E(z; \alpha) = z + \nabla_z (F(z;\alpha) - F(z;0)),
\]
so that $T_{n+1}(x) = E(T_n(x); \alpha_{n+1})$ with $T_0(x)=x$.

The reason for choosing as functional space for the local maps the gradient of the test function is rooted in duality. In particular, under the canonical $L^2$ cost, it follows from the dual of Kantorovich formulation of the OT problem \cite{villani2008optimal} that 
% [XXX]\color{red}(I CAN PROBABLY CITE VILLANI'S TEXTBOOK HERE? Sure!)\color{black}
%
$$ x = y - \nabla_y \psi(y), $$
where the dual function $\psi$ plays the same role as our test function $F(y)$.
% \textcolor{red}{[D: what is this o? A circumflex accent that English borrowed from French. OKAY SO IT"S NOT A TYPO! GOT IT!]} 
At each algorithmic time $n$, we solve the memory-less problem
\[
J_{n} = \min_{\alpha_n} \max_{\beta} \E[c(X, E(T_{n-1}(X),\alpha_n))] + \E[F(E(T_{n-1}(X);\alpha_n);\beta)] - \E[F(Y;\beta)],
\]
or rather perform one ascent-descent step of $J_n$, as in Algorithm \ref{alg:simple} below.

% ALGORITHM: SIMPLE
\begin{algorithm}%[H]
\label{alg:simple}
\SetAlgoLined
Given samples of $X \sim \rho$, $Y \sim \mu$, and $c(\cdot,\cdot)$, perform pre-conditioning \cite{kuang2017preconditioning} on $\rho$, $\mu$. Let $\beta_0=0.$\\
\For{$n\geq 0$, until done}{
    $(d\alpha, d\beta)\leftarrow$ \texttt{ImplicitGradient}$(J_n(\alpha_n,\beta)\rvert_{(0, \beta_n)})$ \cite{essid2019implicit} with additional trust-region cond. \\
    $\left(\alpha_{n+1}, \beta_{n+1}\right)$ $\leftarrow$ $\left(d\alpha, \beta_n + d\beta\right)$ \\
    $T_{n+1}(X) \leftarrow E(T_n(X), \alpha_{n+1})$
}
\Return $(T_1,T_2,\ldots)$, $F$
\caption{Fixed features}
\end{algorithm}
We choose implicit gradient descent for the optimization, but any other minimax algorithm can be used instead. 
The trust region condition we added to the implicit gradient method is that we only allow taking gradient steps with norm up to a prescribed trust region constant $\delta$.
Note that, excluding the transportation cost term, $J_n$ is similar to the objective function of generative adversarial networks (GAN) \cite{goodfellow2014generative}, with $T$ acting as generator and $F$ as the discriminator.

% algorithm
\subsection{General case}
The algorithm above is based on an externally provided set of test functions $\phi_j(y)$. One would like instead to have test functions that adapt to the data, capturing for instance situations when two distributions differ not in their first few moments but in some idiosyncratic, localized details. Proposing a set of test functions that covers all such possibilities is clearly beyond reach, besides almost surely overfitting the data. Instead, we propose a flow-based methodology that builds both the transport maps and the test functions through the composition of elementary maps.

\paragraph{Adversarial test function}
Since $F:\mb{R}^d \to \mb{R}$, we cannot directly define $F$ through a flow. Instead, we write the test function at time $n$ as the difference between the convolution of a kernel $K(b,y)$ and two evolving measures $\nu^{\pm}$:
\[
F_n(y;\beta^+, \beta^-) = \int K(f(b_n^+;\beta^+),y) d\nu^+(b_0^+) - \int K(f(b_n^-;\beta^-),y) d\nu^-(b_0^-).
\]
Here $f(\cdot; \beta)$ is a simple parametric map that reduces to the identity for $\beta=0$. 
The kernel $K:\mb{R}^B\times\mb{R}^d \to \mb{R}$, where $B$ is the dimension of $b$, can be thought of as a similarity function.
Thus we are evolving two initial distributions $\nu^{\pm}(b_0^{\pm})$ through the composition of elementary maps, with $b_{n+1} = f(b_n;\beta_{n+1})$.
The single representer $\beta$ of the prior subsection has been replaced by infinitely many, encompassed by the variables $b^{\pm}$, the \emph{representers} of $F$. The reason to have two such variables is to account for both positive and negative components of $F$.  

Throughout the rest of this paper, we adopt for concreteness $B=d$ and $K$ to be the Gaussian radial basis function kernel %{\textcolor{red}{The one you write below is a Gaussian} I thought Gaussian kernel has another amplitude factor in front of it..?}.
\begin{equation}
K_\sigma(b,y) = \exp\left( -\frac{1}{2\sigma^2} \Vert b-y \Vert^2 \right).\label{eq:gaussian kernel}
\end{equation}
An advantage of making $F$ depend on distributions, is that these can be well-represented through Monte Carlo simulation, using samples $b_i^{\pm}$. Then we have
\[
F_n(y;\beta^+, \beta^-) \approx \frac{1}{N_r}\sum_{i=1}^{N_r} K(f(b_{ni}^+;\beta^+),y) -  K(f(b_{ni}^-;\beta^-),y).
\]
where $N_r$ indicates the number of representers. 
Moreover, as in \cite{tabak2013family}, it is useful to associate to each $b_i$ an adaptive kernel bandwidth $\sigma_i$, larger in areas with small density of $X$ and $Y$, in order to avoid over-fitting. 
The choice of initial distributions for $\nu^{\pm}$ is arbitrary. We use $\nu^+=\nu^-=\mathcal{N}(0,c^2I)$, with a constant $c$ chosen to fit the effective support of $X$ and $Y$. 

\paragraph{Elementary maps}
We consider two options for the elementary maps $f(\cdot,\beta)$ that evolve the representers. 
\begin{itemize}
    \item \textit{Multinomial map}: Just as in the fixed features case, we may consider multinomials. 
    % \textcolor{red}{[QUESTION: DOES LEAVING THE GRADIENT FORM ANY USEFUL FOR BOTH MULTINOMIAL AND RADIAL MAPS? IF NOT I WANT TO REMOVE IT to be consistent with the error function radial map below. I'd leave it as is.]}
    \[
    f(b, \beta) = b + \nabla_b \left( \sum_{j=1}^K \beta_j  \phi_j(b) \right)
    = b + \sum_{j=1}^K \beta_j \nabla_b \phi_j(b)
    \]
    where $\phi(b)$ is a multinomial feature vector of monomials up to degree $D$,
    and $\beta = \langle \beta_1, \ldots, \beta_K \rangle$.
    For $D=2$, the map affects the mean and variance of the evolving measures, while for $D\geq 3$, we have composition of nonlinear transports, a requirement, as the composition of linear maps remains within the linear realm. We usually use $D=3$ to avoid overfitting and achieve computational efficiency. Yet the multinomial map affects the distribution of $b$ globally, resulting often in numerical instability. 
    
    \item \textit{Radial map}: Instead of elementary maps with a global effect on $b$, one can propose local contraction/expansions around a randomly chosen center point $c_0 \in \mb{R}^B$, such as
    \[
    f(b, \beta) = b + \nabla_b \left( \frac{\beta_0}{2} \log(\tau + \Vert b - c_0 \Vert^2) + \beta_1 b \right) = b + \beta_0 \frac{1}{\tau + \Vert b-c_0 \Vert^2}(b-c_0) + \beta_1,
    \]
    where $\beta = \langle \beta_0, \beta_1 \rangle$, with $\beta_0\in\mb{R}$, $\beta_1\in\mb{R}^B$, and $\tau\in\mb{R}^{>0}$ measures the length-scale of the map, which we may choose adaptively as we did for the adaptive bandwidth for the test function.
    Another candidate in this family uses the error function:
    \[
    f(b, \beta) = b + \beta_0 \frac{\textrm{erf}(\Vert b-c_0 \Vert / \tau)}{\Vert b-c_0 \Vert^2}(b-c_0) + \beta_1.
    \]
    Notice that, for $\beta_1=0$, both radial maps have the locality property that $\Vert f(b,\beta) - b \Vert \to 0$ as $\Vert b-c_0\Vert\to\infty.$ 
    An advantage of radial maps is that they have $\mathcal{O}(B)$ many parameters, as compared to the roughly $\mathcal{O}(B^D)$ of multinomial maps,
    while still expressing rich enough functions.
    We choose our center point $c_0$ randomly from the union of sample sets of $T_n(X)$ and $Y$ to explore areas where the push-forward condition may not yet have been met. 
    
    Having one radial map per step usually results in stochastic behavior. To decrease the variance of the resulting oscillations of the objective function, one can pick multiple points randomly at each iteration, using a "mini-batch" version of the algorithm: for $N_c$ indicating number of centers, use
    \[
    f(b, \beta) 
    = b + \frac{1}{N_c}\sum_{i=1}^{N_c} \beta_0^i \frac{1}{\tau + \Vert b-c_i \Vert^2}(b-c_i) + \beta_1^i
    \]
\end{itemize}

% \subsubsection{Similarity Function $K$}

% For our choice, we let $b\sim\sigma$, where $b\in\mb{R}^B=\mb{R}^d$. For some scalars $d_0, d_1 \in\mb{R}$,
% \begin{enumerate}
%     \item \textit{Polynomial Kernel}
%     \al{
%     K(u,v) &= \langle u^T v + d_0 \rangle ^ {d_1}
%     }
%     \item \textit{Radial Basis Kernel}
%     \al{
%     K(u,v) &= d_0 \exp \left( {-d_1 \Vert u-v \Vert^2} \right)
%     }
    
% \end{enumerate}

\paragraph{Transport map}
As in the feature-based procedure, $T$ must follow the gradient of $F$, so we propose a flow-based transport map made up of the composition of elementary maps of the form:
\al{
T_{n+1}(y) &= E_{n+1}(T_n(y); \alpha_{n+1}) \quad \text{with} \quad T_0(y)=y, \\
E_{n+1}(y;\alpha) &= y + \nabla_y \left( F_n(y;\alpha) - F_n(y;0) \right),
}
where $\alpha$ stands for both $\alpha^+$ and  $\alpha^-$, and $E_n(y;0) = y$. 
% \textcolor{brown}{Notice that, at each time $n$, the three elementary maps branch, as they use the same center but different parameters, $\alpha$, $\beta^+$ and $\beta^-$.} 
% \textcolor{red}{[NOTE: INCORRECT. I'm using 4 maps: 2 for $\alpha\pm$, 2 for $\beta\pm$, and each flow has its own center point, but now I think that sharing a center point also makes sense.. In your point of view it makes sense to say that elementary maps branch from the shared center point, but in my view, I think it makes sense to say that representers branch. I'll put it down here.]}
Notice that at each algorithmic time $n$, the representers $b_n$ temporarily branch in the direction of $\alpha$ and perform transporting the mass following the gradient of $\left( F_n(\cdot, \alpha)-F_n(\cdot,0)\right)$, but eventually flow in the direction of $\beta$ before proceeding to the next iteration. The details are described in algorithm \ref{alg:general} below. The objective function we optimize at each iteration is nearly identical to the one in the previous subsection:
\[
J_{n} = \min_{\alpha_{n}} \max_{\beta_{n}} \E[c(X, E_{n}(T_{n-1}(X),\alpha_{n}))] + \E[F_{n-1}(E_{n}(T_{n-1}(X);\alpha_{n});\beta_{n})] - \E[F_{n-1}(Y;\beta_{n})].
\]
% ALGORITHM: GENERAL
\begin{algorithm}%[H]
\label{alg:general}
\SetAlgoLined
Given $X \sim \rho$, $Y \sim \mu$, and  $c(\cdot,\cdot)$, perform pre-conditioning \cite{kuang2017preconditioning} on $\rho$ and $\mu$ and initialize $b_0\sim\nu$.\\
\For{$n\geq 0$, until done}{
    $(\alpha_{n+1}, \beta_{n+1})\leftarrow$ \texttt{ImplicitGradient}$(J_n(\alpha,\beta)\rvert_{(0, 0)})$ \cite{essid2019implicit} with trust-region condition. \\
    $T_{n+1}(X) \leftarrow E_{n+1}(T_n(X), \alpha_{n+1})$\\
    $b_{n+1} \leftarrow f(b_n;\beta_{n+1})$
}
\Return $(T_1,T_2,\ldots)$, $(F_1,F_2,\ldots)$
\caption{General case}
\end{algorithm}

% figure sample format
% \begin{figure}
%   \centering
%   \fbox{\rule[-.5cm]{0cm}{4cm} \rule[-.5cm]{4cm}{0cm}}
%   \caption{Sample figure caption.}
% \end{figure}
\section{Experiments}
\label{section:examples}
We illustrate our algorithm with numerical examples in various space dimensions and conduct empirical analysis on convergence and complexity. 
Throughtout the examples, we use as elementary map the radial maps with error function non-linearity.
Linear radial maps and multinomial maps also work, but nonlinear radial map tends to behave more robustly.
We fix the number of center points at $N_c=1$. Using the mini-batch provides a smoother convergence but runs more slowly.
We use $N_r=100$ representers, with distribution $\nu = \mathcal{N}(0, \frac{1}{4}I)$ for both the positive and negative components. We set the trust region condition scalar $\delta$ to be $0.003$.

\subsection{Optimal map recovery and generative modeling}
We first run a 1-dimensional example, where we can fully visualize the workings of the algorithm. We pick as source a Gaussian, and as target its push-forward by the gradient $T$ of the convex function $\phi(x) = \vert x \vert^{1.5}$, with a weak singularity at $x=0$. 
By Brenier's theorem \cite{villani2008optimal}, we know that $T$ is the unique optimal map.

\begin{figure}
  \centering
  \includegraphics[width=1\linewidth]{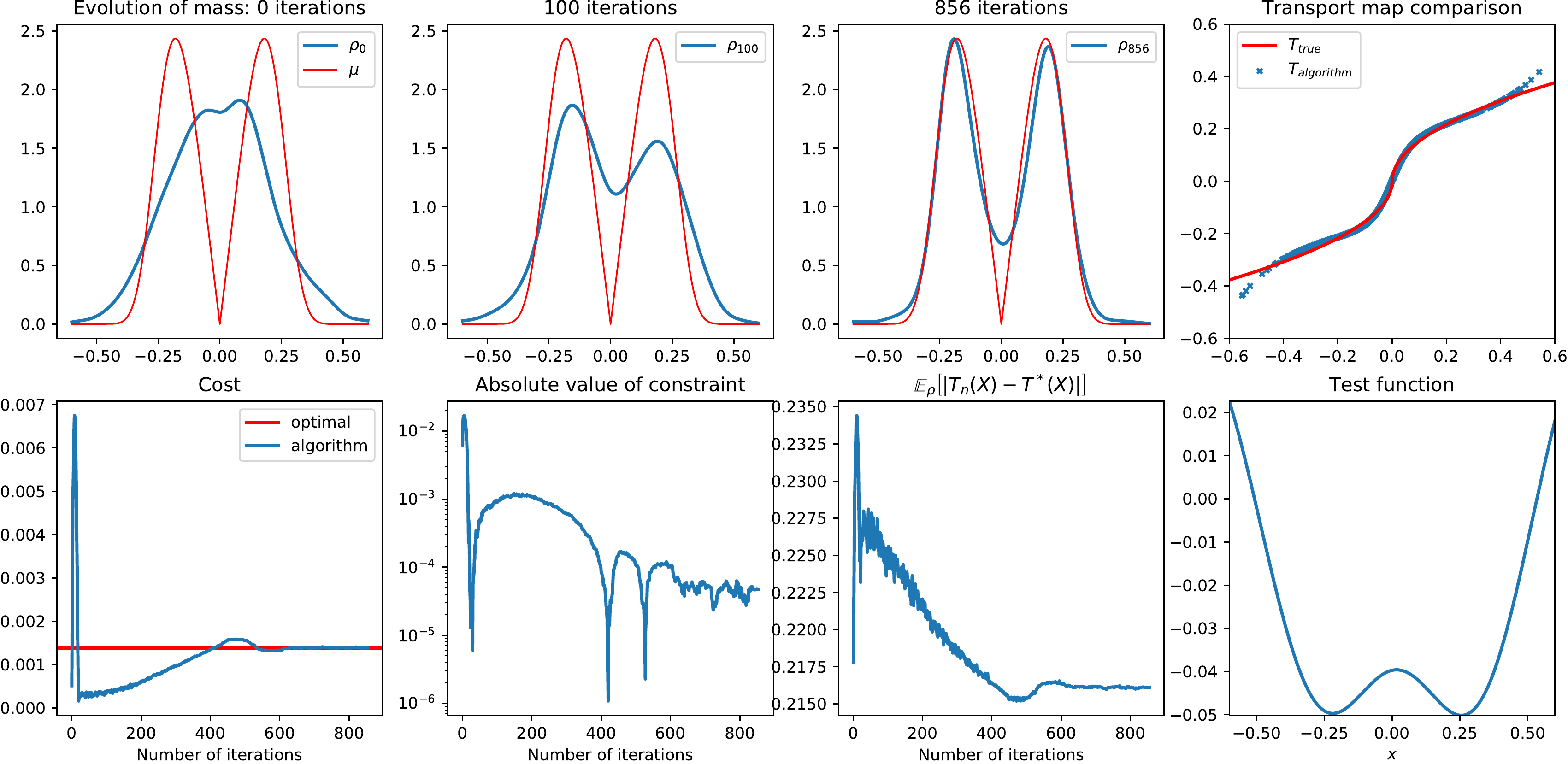}      
  \caption{
  Optimal map recovery. Top row: evolution of mass from Gaussian $\rho$ to bi-modal $T\#\rho$. 
  (The plots display the distributions after the preconditioning step.)
  A Gaussian kernel density estimation is used for the plots at each time, in lieu of an histogram. 
  On the right, the map discovered by the algorithm is plotted for the sample points, along with the true optimal map ${d\phi}/{dx}$. 
  Bottom row: cost, constraint and $L^1$-norm of the error against the number of iterations, and the final test function $F$.}
  \label{fig:recovery}
\end{figure}

We generate two independent batches of 1000 samples each, $X_1, X_2$, from the Gaussian distribution, and apply algorithm \ref{alg:general} with $X_1$ and $T(X_2)$ as data.
Figure \ref{fig:recovery} shows the results. 
Observe that $X_1$ is transported to a bi-modal distribution close to the true target, and that the map found approximates the true $T$ well, except in sample poor areas at the tail of the distribution. 
%At the intermediate step, we see that the mass is not symmetric.
%This is due to the fact that the radial map picks a center point at random.
%Initially, the mass is transported towards one of the mode, and the algorithm fixes from there on for the rest of the time.
We applied the adaptive bandwidth technique \cite{tabak2013family} for test function to assist with the division of mass at the middle.

The test function $F$ that the algorithm finds is also the correct one, with the transport map minus the identity map agreeing with its gradient. 
The cost oscillates around the analytical optimal value, and the constraint oscillates around zero, due to the adversarial nature of the problem formulation.
The evolution of the $\rho$-weighted $L^1$-norm between $T_\textrm{algo}$ and $T_\textrm{true}$,  $\mb{E}_\rho\left[\vert T_{n}(X_1)-T(X_1)\vert\right]$ shows the algorithm transporting mass smoothly from source to target.

In this example, the algorithm pushed forward a Gaussian distribution into a quite different, bi-modal one. Mapping an easily sampleable distribution into another one known only through samples is key to generative models. 
\subsection{Density estimation}
Moving to 2-dimensions, we illustrate our algorithm on the density estimation of a tri-modal Gaussian mixture distribution. In density estimation, we seek a map $T$ that moves a distribution $\rho(x)$ known only through samples to a known distribution $\mu(y)$, such as a Gaussian. 
Then we estimate the source density using the change of variables formula.
\begin{figure}
  \centering
  \includegraphics[width=1\linewidth]{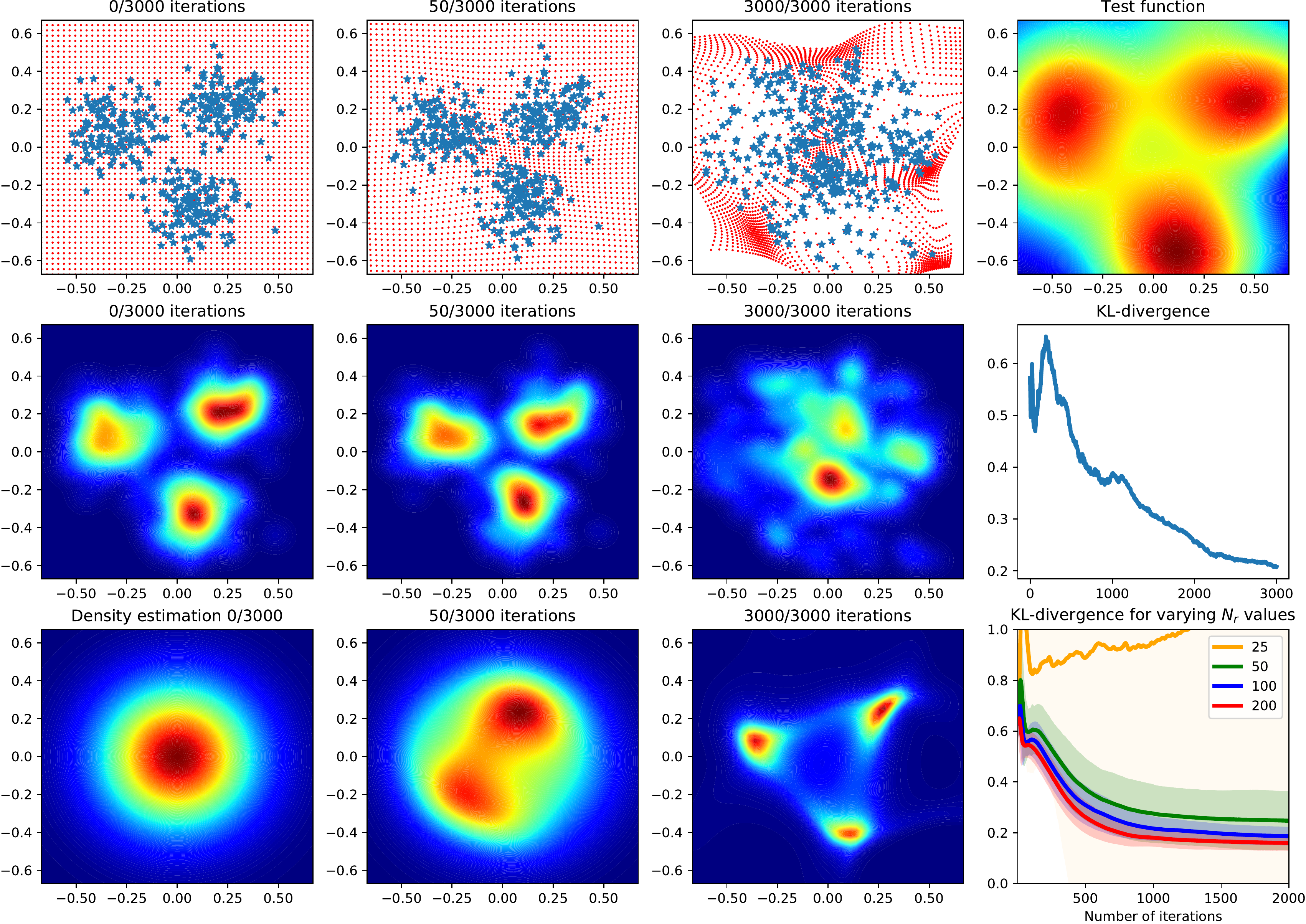}   
  \caption{Density estimation of a tri-modal Gaussian mixture. Top row: scatter plots of the evolving samples  along with the evolution of a passive grid, and final test function $F$. Middle row: Gaussian kernel density estimation on the evolving $T\#\rho$ and KL-divergence against number of iterations. Bottom row: Density estimation for the initial, intermediate and final iterations, and KL-divergence of the algorithm with varying values of the number $N_r$ of representers.}
  \label{fig:density}
\end{figure}

We generate 400 samples from each $X$ and $Y$, and use constant values for $\sigma=0.2$ and $\tau=0.1$, the bandwidths for the test function and the elementary radial maps, respectively. 
The first row in figure \ref{fig:density} shows in blue scatter plots of the evolution of the samples through time, and the corresponding displacement of a passive grid in red. The evolving grid help us visualize the map: following the gradient of the test function, the transport map expands the centers of each mode of the mixture distribution and contracts the interior regions between the clusters in order to push the mass away from the nodes, diffusing it in such a way that the final transportation approximates the target, a uni-modal Gaussian distribution. In addition, the map contracts the exterior regions near the edges of the domain to bring in the extra mass from the tails of the three modes and fill the gaps between them.
From the intermediate step, we see that initially the algorithm mostly moves mass toward $(0,0)$, where the mismatch between $\rho$ and $\mu$ is largest, and then focuses on other areas, as allowed by the explorative nature of the random center points for the radial maps. 

The bottom row shows the density estimation at different iteration steps, 
 $\rho_n(x) := \vert \det{\nabla T(x)} \vert \mu(T(x))$, dividing the uni-modal target into three clear clusters and capturing the valleys between them. 
Since in this synthetic example we know the true probability density $\rho(x),$ we computed the KL-divergence $D_{KL}(\rho \Vert \rho_n)$ via Monte Carlo, and plotted it against the number of iterations. 
Fixing everything else, we ran 100 experiments for varying values of number of representers $N_r$ and computed KL-divergence. The lines in the plot are the means of each individual experiment, and we showed one standard deviation around it. This result indicates that after certain threshold of minimum number of representers, the algorithm performs stably and Monte Carlo estimation of the test function is robust with respect to the estimation through samples. 

\section{Conclusion}
\label{section:conclusion}
\begin{figure}
  \centering
  \includegraphics[width=0.4\linewidth]{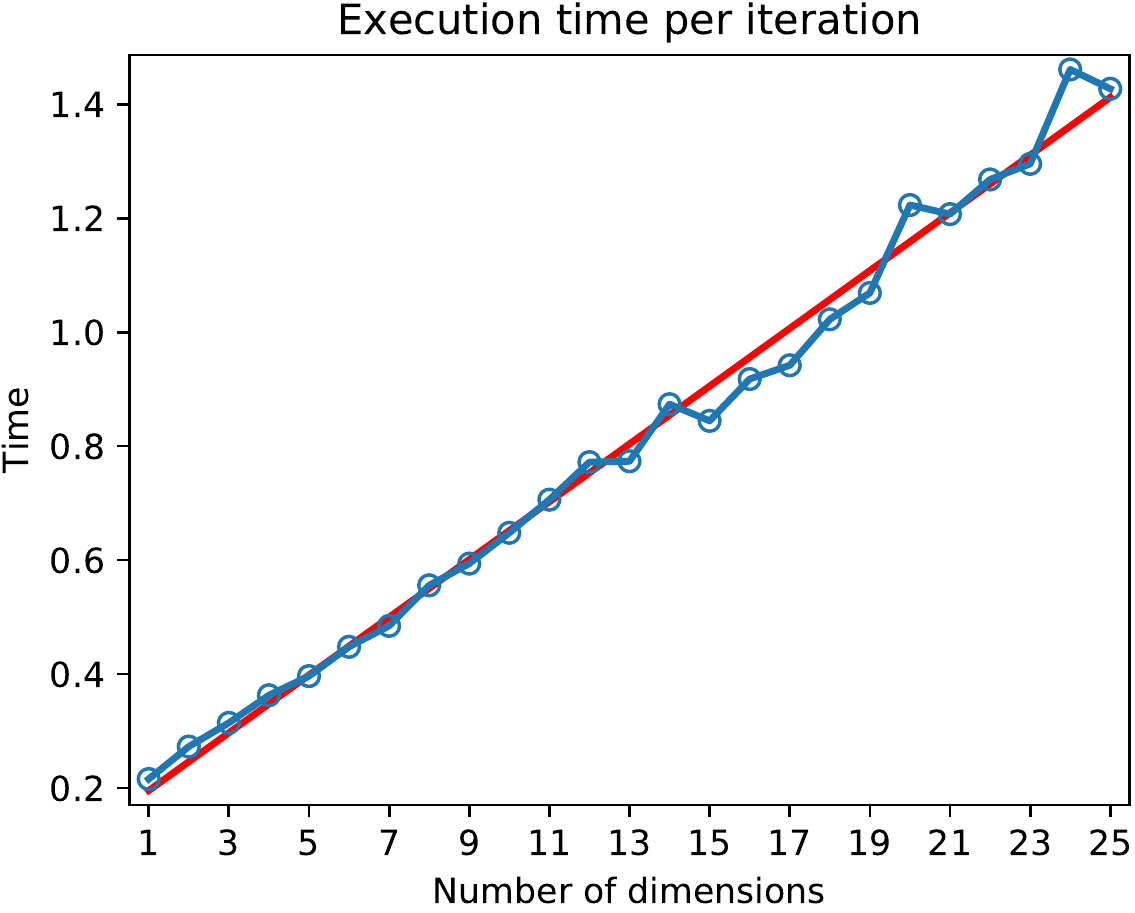}    
  \caption{Execution time per iteration. We run algorithm \ref{alg:general} from the bi-modal Gaussian mixture $(\mathcal{N}(-2\cdot \textbf{1},I_d) + \mathcal{N}(2\cdot\textbf{1},I_d))/2$ to a Gaussian, with all parameters of the algorithm fixed except the dimensionality $d$.}
  \label{fig:time}
\end{figure}

This article introduced a set of novel flow-based function classes for the adversarial formulation of the Monge optimal transport problem and developed an algorithm to solve the problem numerically. 
The adversarial test function proposed is a convolution between a kernel function and two evolving measures of representers, simulated through samples. The representer flow enables us to build up a rich, complex function from composition of the elementary maps in a memory-less fashion. In figure \ref{fig:time}, we demonstrate the running time per iteration against the number of dimensions. We see that the complexity of the  algorithm is indeed linear in space dimension, a potential huge benefit for dealing with high-dimensional problems. 

Sample-based optimal transport has a variety of uses, of which we demonstrated some through 1 and 2-dimensional numerical examples: density estimation, generative models, data normalization, determination of the effect of a treatment (i.e. map discovery). 

Further work in progress includes the application of the methodology to real-life problems, its extension to handle the
Wasserstein barycenter problem, and the development of further improvements, such as automatic determination of the hyperparameters, stochastic descent and online learning.

\section*{Broader Impact}

Advances in the solution to the sample-based optimal transport problem, such as the methodological developments of this article, have a number of potential societal impacts. Typical examples of application include the determination of the effect of a medical treatment, weather forecast, and risk management, through the simulation of future events under a variety of scenarios.  
% \begin{ack}
% Use unnumbered first level headings for the acknowledgments. All acknowledgments
% go at the end of the paper before the list of references. Moreover, you are required to declare 
% funding (financial activities supporting the submitted work) and competing interests (related financial activities outside the submitted work). 
% More information about this disclosure can be found at: \url{https://neurips.cc/Conferences/2020/PaperInformation/FundingDisclosure}.

% Do {\bf not} include this section in the anonymized submission, only in the final paper. You can use the \texttt{ack} environment provided in the style file to autmoatically hide this section in the anonymized submission.
% \end{ack}  % STILL HAVE TO DO THIS!

\bibliography{bib} 
\bibliographystyle{ieeetr}
\end{document}